\documentclass{article}

\usepackage{times}
\usepackage{graphicx} 
\usepackage{subfigure} 
\usepackage{rotating}
\usepackage{multirow}
\usepackage{natbib}
\usepackage{amssymb}
\usepackage{amsmath}
\usepackage{mathtools}
\usepackage{algorithm}
\usepackage{algorithmic}
\usepackage{multirow}
\usepackage{tcolorbox}
\usepackage{tabularx}
\usepackage{array}
\usepackage{colortbl}
\usepackage{fancybox}
\usepackage{tikz}

\tcbuselibrary{skins}

\newcolumntype{Y}{>{\raggedleft\arraybackslash}X}

\usepackage{tikz}

\tcbset{tab3/.style={fonttitle=\bfseries\footnotesize,fontupper=\footnotesize,
colback=yellow!10!white,colframe=red!75!black,colbacktitle=red!40!white,
coltitle=black,center title,freelance}}

\tcbset{tab2/.style={fonttitle=\bfseries\footnotesize,fontupper=\footnotesize,
colback=blue!2!white,colframe=blue!75!black,colbacktitle=red!40!white,
coltitle=black,center title,freelance}}

\usepackage{hyperref}

\usepackage[accepted]{icml2018}

\icmltitlerunning{Automated Clinical Prognostic Modeling via Bayesian Optimization}

\begin{document} 

\twocolumn[
\icmltitle{AutoPrognosis: Automated Clinical Prognostic Modeling\\ via Bayesian Optimization with Structured Kernel Learning}

\icmlsetsymbol{equal}{*}

\begin{icmlauthorlist}
\icmlauthor{Ahmed M. Alaa}{to}
\icmlauthor{Mihaela van der Schaar}{goo}
\end{icmlauthorlist}

\icmlaffiliation{to}{University of California, Los Angeles, USA}
\icmlaffiliation{goo}{Alan Turing Institute, London, UK}


\icmlkeywords{boring formatting information, machine learning, ICML}

\vskip 0.3in
]



\printAffiliationsAndNotice{\icmlEqualContribution} 

\begin{abstract}
Clinical prognostic models derived from large-scale healthcare data can inform critical diagnostic and therapeutic decisions. To enable off-the-shelf usage of machine learning (ML) in prognostic research, we developed {\footnotesize \textsc{AutoPrognosis}}: a system for automating the design of predictive modeling {\it pipelines} tailored for clinical prognosis. {\footnotesize \textsc{AutoPrognosis}} optimizes ensembles of pipeline configurations efficiently using a novel batched Bayesian optimization (BO) algorithm that learns a low-dimensional decomposition of the pipelines' high-dimensional hyperparameter space in concurrence with the BO procedure. This is achieved by modeling the pipelines' performances as a black-box function with a Gaussian process prior, and modeling the ``similarities" between the pipelines' baseline algorithms via a sparse additive kernel with a Dirichlet prior. {\it Meta-learning} is used to warmstart BO with external data from ``similar" patient cohorts by calibrating the priors using an algorithm that mimics the empirical Bayes method. The system automatically explains its predictions by presenting the clinicians with logical {\it association rules} that link patients' features to predicted risk strata. We demonstrate the utility of {\footnotesize \textsc{AutoPrognosis}} using 10 major patient cohorts representing various aspects of cardiovascular patient care. 
\end{abstract}

\section{Introduction} 
\label{sec1} 
In clinical medicine, {\it prognosis} refers to the risk of future health outcomes in patients with given features. Prognostic~research~aims~at building actionable predictive models that can inform clinicians about future course of patients' clinical conditions in order to guide screening and therapeutic decisions. With the recent abundance of data linkages, electronic health records, and bio-repositories, clinical researchers have become aware that the value conferred by big, heterogeneous clinical data can only be realized with prognostic models based on flexible machine learning (ML) approaches. There is, however, a concerning gap between the potential and actual utilization of ML in prognostic research; the reason being that clinicians with no expertise in data science find it hard to manually design and tune ML pipelines \cite{luo2017automating}.

\begin{figure}[t]
\centering
\includegraphics[width=3.25in]{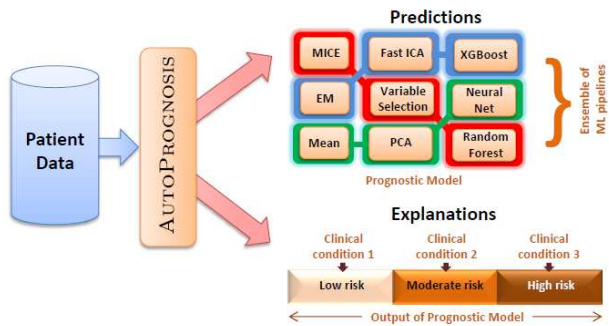}
\caption{\small Illustration for exemplary outputs of {\footnotesize \textsc{AutoPrognosis}}.}
\label{Fig0}
\end{figure}

To fill this gap, we developed {\footnotesize \textsc{AutoPrognosis}}, an automated ML (AutoML) framework tailored for clinical prognostic modeling. {\footnotesize \textsc{AutoPrognosis}} takes as an input data from a patient cohort, and uses such data to automatically configure ML {\it pipelines}. Every ML pipeline comprises all stages of prognostic modeling: missing data imputation, feature preprocessing, prediction, and calibration. The system handles different types of clinical data, including longitudinal and survival (time-to-event) data, and automatically {\it explains} its predictions to the clinicians via an ``interpreter" module which outputs clinically interpretable associations between patients' features and predicted risk strata. An overview of the system is provided in Figure \ref{Fig0}. 

The core component of {\footnotesize \textsc{AutoPrognosis}} is an algorithm for configuring ML pipelines using Bayesian optimization (BO) \cite{snoek2012practical}. Our BO algorithm models the pipelines' performances as a black-box function, the input to which is a ``pipeline configuration", i.e. a selection of algorithms and hyperparameter settings, and the output of which is the performance (predictive accuracy) achieved by such a configuration. We implement BO with a {\it Gaussian process} (GP) prior on the black-box function. To deal with the high-dimensionality of the pipeline configuration space, we capitalize on the fact that for a given dataset, {\bf the performance of one ML algorithm may not be correlated with that of another algorithm}. For instance, it may be the case that the observed empirical performance of {\it logistic regression} on a given dataset does not tell us much information about how a {\it neural network} would perform on the same dataset. In such a case, both algorithms should not share the same GP prior, but should rather be modeled independently. Our BO {\bf learns} such a decomposition of algorithms from data in order to break down the high-dimensional optimization problem into a set of lower-dimensional sub-problems. We model the decomposition of algorithms via an additive kernel with a Dirichlet prior on its structure, and learn the decomposition from data in concurrence with the BO iterations. We also propose a batched (parallelized) version of the BO procedure, along with a computationally efficient algorithm for maximizing the BO acquisition function. 
 
{\footnotesize \textsc{AutoPrognosis}} follows a principled Bayesian approach in all of its components. The system implements post-hoc construction of pipeline ensembles via {\it Bayesian model averaging}, and implements a {\it meta-learning} algorithm that utilizes data from external cohorts of ``similar" patients using an {\it empirical Bayes} method. In order to resolve the tension between accuracy and interpretability, which is crucial for clinical decision-making \cite{cabitza2017unintended}, the system presents the clinicians with a rule-based approximation for the learned ML pipeline by mining for logical associations between patients' features and the model's predicted risk strata using a {\it Bayesian associative classifier} \cite{agrawal1993mining,kruschke2008bayesian}. 

We conclude the paper by conducting a set of experiments on multiple patient cohorts representing various aspects of cardiovascular patient care, and show that prognostic models learned by {\footnotesize \textsc{AutoPrognosis}} outperform widely used clinical risk scores and existing AutoML frameworks.\\

{\bf Related work:} To the best of our knowledge, none of the existing AutoML frameworks, such as {\footnotesize \textsc{Auto-Weka}} \cite{kotthoff2016auto}, {\footnotesize \textsc{Auto-sklearn}} \cite{feurer2015efficient}, and {\footnotesize \textsc{TPOT}} \cite{olson2016tpot} use principled GP-based BO to configure ML pipelines. All of the existing frameworks model the sparsity of the pipelines' hyperparameter space via frequentist tree-based structures. Both {\footnotesize \textsc{Auto-Weka}} and {\footnotesize \textsc{Auto-sklearn}} use BO, but through tree-based heuristics, such as random forest models and tree Parzen estimators, whereas TPOT uses a tree-based genetic programming algorithm. Previous works have refrained from using principled GP-based BO because of its statistical and computational complexity in high-dimensional hyperparameter spaces. Our algorithm makes principled, high-dimensional GP-based BO possible by {\bf learning} a sparse additive kernel decomposition for the GP prior. This approach confers many advantages as it captures the uncertainty about the sparsity structure of the GP prior, and allows for principled approaches for (Bayesian) meta-learning and ensemble construction that are organically connected to the BO procedure. In Section \ref{sec4}, we compare the performance of {\footnotesize \textsc{AutoPrognosis}} with that of {\footnotesize \textsc{Auto-Weka}}, {\footnotesize \textsc{Auto-sklearn}}, and {\footnotesize \textsc{TPOT}}, demonstrating the superiority of our algorithm.    

Various previous works have addressed the problem of high-dimensional GP-based BO. \cite{wang2013bayesian} identifies a low-dimensional effective subspace for the black-box function via random embedding. However, in the AutoML setup, this approach cannot incorporate our prior knowledge about dependencies between the different hyperparameters (we know the sets of hyperparameters that are ``activated" upon selecting an algorithm \cite{hutter2011sequential}). This prior knowledge was captured by the {\it Arc-kernel} proposed in \cite{swersky2014raiders}, and similarly in \cite{jenatton2017bayesian}, where a BO algorithm for domains with tree-structured dependencies was proposed. Unfortunately, both methods require full prior knowledge of the dependencies between the hyperparameters, and hence cannot be used when jointly configuring hyperparameters across multiple algorithms, since the correlations of the performances of different algorithms are not known a priori. \cite{bergstra2011implementations} proposed a na\"ive approach that defines an independent GP for every set of hyperparameters that belong to the same algorithm. Since it does not share any information between the different algorithms, this approach would require trying all combinations of algorithms in a pipeline exhaustively. (In our system, there are 4,800 possible pipelines.) Our model solves the problems above via a {\bf data-driven} kernel decomposition, through which only relevant groups of hyperparameters share a common GP prior, thereby balancing the trade-off between ``information sharing" among hyperparameters and statistical efficiency. 

\begin{figure*}[h]
\centering
\includegraphics[width=6in]{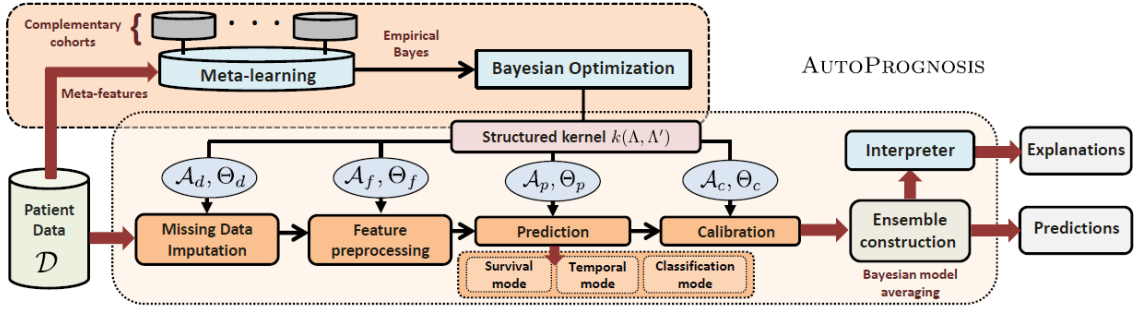}
\caption{\small A schematic depiction of {\footnotesize \textsc{AutoPrognosis}}. Every ML pipeline comprises imputation, feature processing, prediction, and calibration algorithms. The ensemble construction and interpreter modules are included in the system as post-processing steps.}
\label{fig1}
\end{figure*}

\section{\textsc{AutoPrognosis}: A Practical System for Automated Clinical Prognostic Modeling}
\label{sec2}
Consider a dataset $\mathcal{D} = \{(x_i,y_i)\}^n_{i=1}$ for a cohort of $n$ patients, with $x_i$ being patient $i$'s features, and $y_i$ being the patient's clinical endpoint. {\footnotesize \textsc{AutoPrognosis}} takes $\mathcal{D}$ as an input, and outputs an automatically configured {\it prognostic model} which predicts the patients' risks, along with ``explanations" for the predicted risk strata. This Section provides an overview of the components of {\footnotesize \textsc{AutoPrognosis}}; a schematic depiction of the system is shown in Figure \ref{fig1}.  

The core component of {\footnotesize \textsc{AutoPrognosis}} is an algorithm that automatically configures ML pipelines, where every pipeline comprises algorithms for missing data imputation (${\tiny \square}$), feature preprocessing (${\tiny \clubsuit}$), prediction ($\bullet$), and calibration (${\tiny \bigstar}$). Table \ref{Table1} lists the baseline algorithms adopted by the system in all the stages of a pipeline. The imputation and calibration stages are particularly important for clinical prognostic modeling \cite{blaha2016critical}, and are not supported in existing AutoML frameworks. The total number of hyperparameters in {\footnotesize \textsc{AutoPrognosis}} is 106, which is less than those of {\footnotesize \textsc{Auto-Weka}} (786) and {\footnotesize \textsc{Auto-sklearn}} (110). The pipeline configuration algorithm uses {\bf Bayesian optimization} to estimate the performance of different pipeline configurations in a {\bf scalable} fashion by learning a {\bf structured kernel decomposition} that identifies algorithms with {\it correlated} performance. Details of the Bayesian optimization algorithm are provided in Sections \ref{sec3} and \ref{sec4}.

In order to cope with the diverse nature of clinical data and health outcomes, {\footnotesize \textsc{AutoPrognosis}} pipelines are enriched with three modes of operation: {\bf (a) classification mode}, {\bf (b) temporal mode}, and {\bf (c) survival mode}. The {\bf classification mode} handles datasets with binary clinical outcomes \cite{yoon2017personalized}. In this mode, the baseline predictive models include all algorithms in the \texttt{scikit-learn} library \cite{pedregosa2011scikit}, in addition to other powerful algorithms, such as XGBoost \cite{chen2016xgboost}. The {\bf temporal mode} handles longitudinal and time series data \cite{alaa2017learning} by applying the classification algorithms above on data residing in a sliding window within the time series, which we parametrize by the sequence time \cite{hripcsak2015parameterizing}. The {\bf survival mode} handles {\it time-to-event} data, and involves all the classification algorithms above, in addition to survival models such as Cox proportional hazards model and survival forests \cite{ishwaran2008random}, and models for multiple competing risks \cite{fine1999proportional}.

The {\bf meta-learning} module is a pre-processing step that is used to warmstart BO using data from external cohorts, whereas the {\bf ensemble construction} and {\bf interpreter} modules post-process the BO outputs. All of the three module run with a relatively low computational burden. Details of the three modules are provided in Sections \ref{secccc3} and \ref{sec4}. 
       
\begin{table*}[!htbp]
\centering
\begin{tcolorbox}[tab3,tabularx={l||lllll}]
{\bf {\footnotesize Pipeline Stage}} & & & {\footnotesize {\bf Algorithms}} & &  \\
\hline
\hline
{\tiny \bf $\square$} {\footnotesize Data Imputation} & {\tiny $\square$} {\footnotesize missForest {\bf \textcolor{brown}{(2)}}} & {\tiny $\square$} {\footnotesize Median {\bf \textcolor{brown}{(0)}}} & {\tiny $\square$} {\footnotesize Most-frequent {\bf \textcolor{brown}{(0)}}} & {\tiny $\square$} {\footnotesize Mean {\bf \textcolor{brown}{(0)}}} & {\tiny $\square$} {\footnotesize EM {\bf \textcolor{brown}{(1)}}}  \\
& {\tiny $\square$} {\footnotesize Matrix completion {\bf \textcolor{brown}{(2)}}} & {\tiny $\square$} {\footnotesize MICE {\bf \textcolor{brown}{(1)}}} & {\tiny $\square$} {\footnotesize None {\bf \textcolor{brown}{(0)}}} & &  \\
\hline \hline
{\tiny $\clubsuit$} {\footnotesize Feature process.} & {\tiny $\clubsuit$} {Feature agglo. {\bf \textcolor{brown}{(4)}}} & {\tiny $\clubsuit$} {Kernel PCA {\bf \textcolor{brown}{(5)}}} & {\tiny $\clubsuit$} {Polynomial {\bf \textcolor{brown}{(3)}}} & {\tiny $\clubsuit$} {Fast ICA {\bf \textcolor{brown}{(4)}}} & {\tiny $\clubsuit$} {PCA {\bf \textcolor{brown}{(2)}}} \\
& {\tiny $\clubsuit$} {R. kitchen sinks {\bf \textcolor{brown}{(2)}}} & {\tiny $\clubsuit$} {Nystroem {\bf \textcolor{brown}{(5)}}} & {\tiny $\clubsuit$} {Linear SVM {\bf \textcolor{brown}{(3)}}} & {\tiny $\clubsuit$} {Select Rates {\bf \textcolor{brown}{(3)}}} & {\tiny $\clubsuit$} {None {\bf \textcolor{brown}{(0)}}} \\
\hline \hline
{\footnotesize $\bullet$ Prediction} & {\footnotesize $\bullet$ Bernoulli NB {\bf \textcolor{brown}{(2)}}} & {\footnotesize $\bullet$ AdaBoost {\bf \textcolor{brown}{(4)}}} & {\footnotesize $\bullet$ Decision Tree {\bf \textcolor{brown}{(4)}}} & {\footnotesize $\bullet$ Grad. Boost. {\bf \textcolor{brown}{(6)}}} & {\footnotesize $\bullet$ LDA {\bf \textcolor{brown}{(4)}}}\\
 & {\footnotesize $\bullet$ Gaussian NB {\bf \textcolor{brown}{(0)}}} & {\footnotesize $\bullet$ XGBoost {\bf \textcolor{brown}{(5)}}} & {\footnotesize $\bullet$ Extr. R. Trees {\bf \textcolor{brown}{(5)}}} & {\footnotesize $\bullet$ Light GBM {\bf \textcolor{brown}{(5)}}} & {\footnotesize $\bullet$ L. SVM {\bf \textcolor{brown}{(4)}}}\\
 & {\footnotesize $\bullet$ Multinomial NB {\bf \textcolor{brown}{(2)}}} & {\footnotesize $\bullet$ R. Forest {\bf \textcolor{brown}{(5)}}} & {\footnotesize $\bullet$ Neural Net. {\bf \textcolor{brown}{(5)}}} & {\footnotesize $\bullet$ Log. Reg. {\bf \textcolor{brown}{(0)}}} & {\footnotesize $\bullet$ GP {\bf \textcolor{brown}{(3)}}}\\
 & {\footnotesize $\bullet$ Ridge Class. {\bf \textcolor{brown}{(1)}}} & {\footnotesize $\bullet$ Bagging {\bf \textcolor{brown}{(4)}}} & {\footnotesize $\bullet$ $k$-NN {\bf \textcolor{brown}{(1)}}} & {\footnotesize $\bullet$ Surv. Forest {\bf \textcolor{brown}{(5)}}} & {\footnotesize $\bullet$ Cox Reg. {\bf \textcolor{brown}{(0)}}} \\
\hline \hline
{\tiny $\bigstar$} {\footnotesize Calibration} & {\tiny $\bigstar$} {\footnotesize Sigmoid {\bf \textcolor{brown}{(0)}}} & {\tiny $\bigstar$} {\footnotesize Isotonic} {\bf \textcolor{brown}{(0)}} & {\tiny $\bigstar$} {\footnotesize None {\bf \textcolor{brown}{(0)}}} \\
\end{tcolorbox}
\caption{List of algorithms included in every stage of the pipeline. Numbers in brackets correspond to the number of hyperparameters.}
\label{Table1}
\end{table*}

\section{Pipeline Configuration via Bayesian Optimization with Structured Kernels}
\label{sec3}
Let $(\mathcal{A}_d,\mathcal{A}_f,\mathcal{A}_p,\mathcal{A}_c)$ be the sets of all missing data imputation, feature processing, prediction, and calibration algorithms considered in {\footnotesize \textsc{AutoPrognosis}} (Table \ref{Table1}), respectively. A {\bf pipeline} $P$ is a tuple of the form: 
\begin{align}
\Ovalbox{\mbox{\footnotesize $P = (A_d,A_f,A_p,A_c)$}} \nonumber
\end{align}
where $A_v \in \mathcal{A}_v,\, \forall v \in \mbox{\footnotesize $\{d,f,p,c\}$}$. The space of all possible pipelines is given by $\mbox{\footnotesize $\mathcal{P} = \mathcal{A}_d \times \mathcal{A}_f \times \mathcal{A}_p \times \mathcal{A}_c$}$. Thus, a pipeline is a selection of algorithms from the elements of Table \ref{Table1}. An exemplary pipeline can be specified as follows: $\mbox{\footnotesize \bf $P = \{\mbox{\footnotesize \bf MICE}, \mbox{\footnotesize \bf PCA}, \mbox{\footnotesize \bf Random Forest},\mbox{\footnotesize \bf Sigmoid}\}$}$. The total number of pipelines in {\footnotesize \textsc{AutoPrognosis}} is $|\mathcal{P}| = 4,800$.

The specification of a {\bf pipeline configuration} is completed by determining the hyperparameters of its constituting algorithms. The space of hyperparameter configurations for a pipeline is $\Theta = \Theta_d \times \Theta_f \times \Theta_p \times \Theta_c$, where $\Theta_v = \cup_{a}\Theta^{a}_v$, for $v \in \mbox{\footnotesize $\{d,f,p,c\}$},$ with $\Theta^{a}_v$ being the space of hyperparameters associated with the $a^{th}$ algorithm in $\mathcal{A}_v$. Thus, a pipeline configuration $P_\theta \in \mathcal{P}_{\Theta}$ is a selection of algorithms $P \in \mathcal{P}$, and hyperparameter settings $\theta \in \Theta$; $\mathcal{P}_{\Theta}$ is the space of all possible pipeline configurations. 

\subsection{The Pipeline Selection \& Configuration Problem}

The main goal of {\footnotesize \textsc{AutoPrognosis}} is to identify the best pipeline configuration $P^*_{\theta^*} \in \mathcal{P}_{\Theta}$ for a given patient cohort $\mathcal{D}$ via $J$-fold cross-validation as follows: 
\begin{align}
\mbox{\footnotesize $P^*_{\theta^*} \in \arg \max_{P_\theta \in \mathcal{P}_\Theta}\, \frac{1}{J}\sum_{i=1}^{J}\mathcal{L}(P_\theta;\mathcal{D}^{(i)}_{\mbox{\tiny train}},\mathcal{D}^{(i)}_{\tiny \mbox{valid}})$},
\label{eq1}
\end{align}
where $\mathcal{L}$ is a given accuracy metric (AUC-ROC, c-index, etc), $\mathcal{D}^{(i)}_{\mbox{\tiny train}}$ and $\mathcal{D}^{(i)}_{\tiny \mbox{valid}}$ are training and validation splits of $\mathcal{D}$ in the $i^{th}$ fold. The optimization problem in (\ref{eq1}) is dubbed the {\it Pipeline Selection and Configuration Problem} (PSCP). The PSCP can be thought of as a generalization for the {\it combined algorithm selection and hyperparameter optimization} (CASH) problem in \cite{feurer2015efficient,kotthoff2016auto}, which maximizes an objective with respect to selections of single algorithms from the set $\mathcal{A}_p$, rather than selections of full-fledged pipelines from $\mathcal{P}_{\Theta}$.

\subsection{Solving the PSCP via Bayesian Optimization}

The objective in (\ref{eq1}) has no analytic form, and hence we treat the PSCP as a {\it black-box} optimization problem. In particular, we assume that $\mbox{\footnotesize $\frac{1}{J}\sum_{i=1}^{J}\mathcal{L}(P_\theta;\mathcal{D}^{(i)}_{\mbox{\tiny train}},\mathcal{D}^{(i)}_{\tiny \mbox{valid}})$}$ is a noisy version of a black-box function $\mbox{\footnotesize $f: {\bf \Lambda} \to \mathbb{R}$}$, were $\mbox{\footnotesize ${\bf \Lambda} = \Theta \times \mathcal{P}$}$, and use BO to search for the pipeline configuration $P^*_{\theta^*}$ that maximizes the black-box function $f(.)$ \cite{snoek2012practical}. The BO algorithm specifies a Gaussian process (GP) prior on $f(.)$ as follows:
\begin{align}
\mbox{\footnotesize $f \sim \mathcal{GP}(\mu(\Lambda),k(\Lambda,\Lambda^{\prime}))$},
\label{eq10X}
\end{align}
where $\mu(\Lambda)$ is the {\it mean function}, encoding the expected performance of different pipeline, and $k(\Lambda,\Lambda^{\prime})$ is the {\it covariance kernel} \cite{rasmussen2006gaussian}, encoding the similarity between the different pipelines.

\begin{figure*}[t]
\centering
\includegraphics[width=6in]{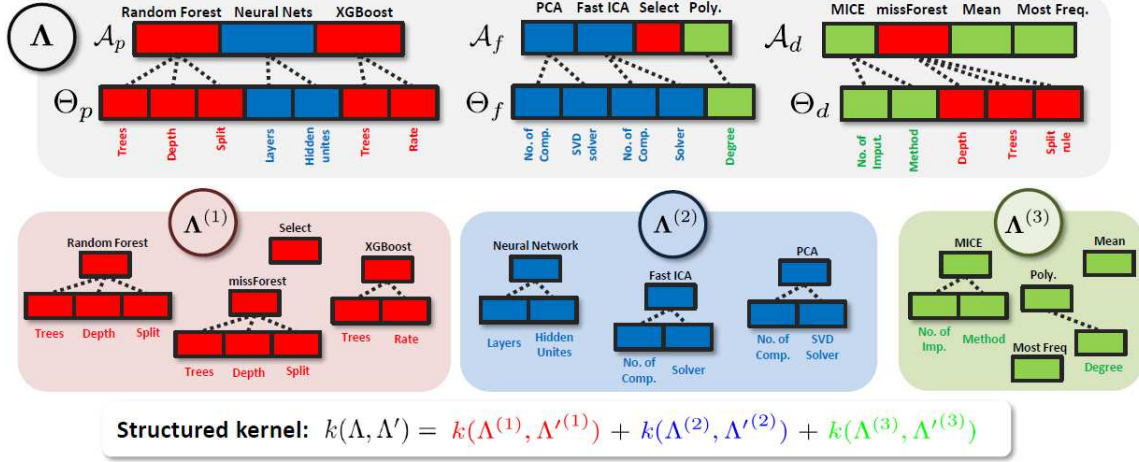}
\caption{\small Illustration for a exemplary subspace decomposition $\{{\bf \Lambda}^{(m)}\}^3_{m=1}$.}
\label{fig2}
\end{figure*}

\subsection{Bayesian Optimization via Structured Kernels}  
The function $f$ is defined over the \mbox{\footnotesize $D$}-dimensional space $\mbox{\footnotesize ${\bf \Lambda}$}$, where $\mbox{\footnotesize $D = \mbox{dim}({\bf \Lambda})$}$ is given by 
\begin{align}
\mbox{\footnotesize $D = \mbox{\small dim}(\mathcal{P}) + {\textstyle \sum}_{v \in \{d,f,p,c\}} {\textstyle \sum}_{a \in \mathcal{A}_v} \mbox{\small dim}(\Theta^a_v).$}
\label{eq20X}
\end{align} 
In {\footnotesize \textsc{AutoPrognosis}}, the domain $\mbox{\footnotesize ${\bf \Lambda}$}$ is high-dimensional, with $\mbox{\footnotesize $D = 106$}$. (The dimensionality of $\mbox{\footnotesize ${\bf \Lambda}$}$ can be calculated by summing up the number of pipeline stages and the number of hyperparameters in Table \ref{Table1}.) High-dimensionality renders standard GP-based BO infeasible as both the sample complexity of nonparametric estimation and the computational complexity of maximizing the acquisition function are exponential in $\mbox{\footnotesize $D$}$ \cite{gyorfi2006distribution,kandasamy2015high}. For this reason, existing AutoML frameworks have refrained from using GP priors, and relied instead on scalable tree-based heuristics \cite{feurer2015efficient,kotthoff2016auto}. Despite its superior performance, recent empirical findings have shown that plain-vanilla GP-based BO is feasible only for problems with $\mbox{\footnotesize $D \leq 10$}$ \cite{wang2013bayesian}. Thus, the deployment of GP-based BO has been limited to hyperparameter optimization for single, pre-defined ML models via tools such as \texttt{Google}'s \texttt{Visier} and \texttt{HyperTune} \cite{golovin2017google}. {\footnotesize \textsc{AutoPrognosis}} overcomes this challenge by leveraging the structure of the PSCP problem as we show in what follows. 

\subsubsection{The Structure of the PSCP Problem} 
The key idea of our BO algorithm is that for a given dataset, {\bf the performance of a given group of algorithms may not be \textit{informative} of the performance of another group of algorithms}. Since the kernel \mbox{\footnotesize $k(\Lambda,\Lambda^{\prime})$} encodes the correlations between the performances of the different pipeline configurations, the underlying ``informativeness" structure that relates the different hyperparameters can be expressed via the following {\bf sparse additive kernel decomposition}:  
\begin{align}
\mbox{\footnotesize $k(\Lambda,\Lambda^{\prime}) = {\textstyle \sum}_{m=1}^{M} k_m(\Lambda^{(m)},{\Lambda^{\prime}}^{(m)}),$}
\label{eq3}
\end{align}
where \mbox{\footnotesize $\Lambda^{(m)} \in {\bf \Lambda}^{(m)},\forall m \in \{1,.\,.\,.,M\},$} with \mbox{\footnotesize $\{{\bf \Lambda}^{(m)}\}_m$} being a set of {\it disjoint} subspaces of \mbox{\footnotesize ${\bf \Lambda}$}. (That is, \mbox{\footnotesize $\cup_m \Lambda^{(m)} = {\bf \Lambda}$}, and \mbox{\footnotesize $\Lambda^{(m)} \cap \Lambda^{(m^{\prime})} = \emptyset$}.) The subspaces are assigned mutually exclusive subsets of the dimensions of \mbox{\footnotesize ${\bf \Lambda}$}, so that \mbox{\footnotesize ${\textstyle \sum}_{m} \mbox{\small dim}(\Lambda^{(m)}) = D$}. The structure of the kernel in (\ref{eq3}) is {\bf unknown} a priori, and needs to be {\bf learned from data}. The kernel decomposition breaks down $f$ as follows:
\begin{align}
\mbox{\footnotesize $f(\Lambda) = {\textstyle \sum}_{m=1}^{M} f_m(\Lambda^{(m)}).$}
\label{eq2}
\end{align}
The additively sparse structure in (\ref{eq3}) gives rise to a statistically efficient BO procedure. That is, if $f$ is $\gamma$-smooth, then our additive kernels reduce {\bf sample complexity} from \mbox{\footnotesize $\boldsymbol{O(n^{\frac{-\gamma}{2\gamma + D}})}$} to \mbox{\footnotesize $\boldsymbol{O(n^{\frac{-\gamma}{2\gamma + D_m}})}$}, where $D_m$ is the maximum number of dimensions in any subspace \cite{raskutti2009lower,yang2015minimax}. (Similar improvements hold for the cumulative regret \cite{kandasamy2015high}.)

Each subspace $\mbox{\footnotesize ${\bf \Lambda}^{(m)} \subset {\bf \Lambda}$}$ contains the hyperparameters of algorithms with correlated performances, whereas algorithms residing in two different subspaces $\mbox{\footnotesize ${\bf \Lambda}^{(m)}$}$ and $\mbox{\footnotesize ${\bf \Lambda}^{(m^{\prime})}$}$ have uncorrelated performances. Since a hyperparameter in \mbox{\footnotesize $\Theta$} is only {\it relevant} to $f(.)$ when the corresponding algorithm in \mbox{\footnotesize $\mathcal{P}$} is selected \cite{hutter2009paramils}, then the decomposition \mbox{\footnotesize $\{{\bf \Lambda}^{(m)}\}_m$} must ensure that all the hyperparameters of the same algorithm are bundled together in the same subspace. This a priori knowledge about the ``conditional relevance" of the dimensions of $\mbox{\footnotesize ${\bf \Lambda}$}$ makes it easier to learn the kernel decomposition from data. Figure \ref{fig2} provides an illustration for an exemplary subspace decomposition for the hyperparameters of a set of prediction, feature processing and imputation algorithms. Since the structured kernel in (\ref{eq3}) is not fully specified a priori, we propose an algorithm to learn it from the data in the next Section.  

\subsubsection{Structured Kernel Learning} 
{\footnotesize \textsc{AutoPrognosis}} uses a Bayesian approach to learn the subspace decomposition \mbox{\footnotesize $\{{\bf \Lambda}^{(m)}\}_m$} in concurrence with the BO procedure, where the following Dirichlet-Multinomial prior is placed on the structured kernel \cite{wang2017batched}: 
\begin{align}
\mbox{$\alpha \sim \mbox{Dirichlet}(M,\gamma),\,\, z_{v,a} \sim \mbox{Multi}(\alpha),$} 
\label{eq4}
\end{align}
\mbox{\footnotesize $\forall a \in \mathcal{A}_v, {\tiny v \in \{d,f,p,c\}}$}, where \mbox{\footnotesize $\gamma = \{\gamma_m\}_m$} is the parameter of a Dirichlet prior, \mbox{\footnotesize $\alpha = \{\alpha_m\}_m$} are the Multinomial mixing proportions, and \mbox{\footnotesize $z_{v,a}$} is an indicator variable that determines the subspace to which the $a^{th}$ algorithm in \mbox{\footnotesize $\mathcal{A}_v$} belongs. The kernel decomposition in (\ref{eq3}) is learned by updating the posterior distribution of \mbox{\footnotesize $\{{\bf \Lambda}^{(m)}\}_m$} in every iteration of the BO procedure. The posterior distribution over the variables \mbox{\footnotesize $\{z_{v,a}\}_{v,a}$} and \mbox{\footnotesize $\alpha$} is given by: 
\begin{align}
\mbox{\footnotesize $\mathbb{P}(z, \alpha\,|\,\mathcal{H}_t,\gamma) \propto \mathbb{P}(\mathcal{H}_t\,|\,z) \, \mathbb{P}(z\,|\alpha) \, \mathbb{P}(\alpha,\gamma),$}
\label{eq5}
\end{align}
where \mbox{\footnotesize $z = \{z_{v,a}: \forall a \in \mathcal{A}_v, \forall v \in \{d,f,p,c\}\}$}, and \mbox{\footnotesize $\mathcal{H}_t$} is the history of evaluations of the black-box function up to iteration $t$. Since the variables \mbox{\footnotesize $\{z_{v,a}\}_{v,a}$} are sufficient statistics for the subspace decomposition, the posterior over \mbox{\footnotesize $\{{\bf \Lambda}^{(m)}\}_m$} is fully specified by (\ref{eq5}) marginalized over \mbox{\footnotesize $\alpha$}, which can be evaluated using Gibbs sampling as follows:
\begin{align}
\mbox{\footnotesize $\mathbb{P}(z_{v,a} = m\,|\,z/\{z_{v,a}\},\mathcal{H}_t) \propto \mathbb{P}(\mathcal{H}_t\,|\,z) \, (|\mathcal{A}_v^{(m)}|+\gamma_{m}),$} \nonumber 
\end{align}
where \mbox{\footnotesize $\mathbb{P}(\mathcal{H}_t\,|\,z)$} is the GP likelihood under the kernel induced by \mbox{\footnotesize $z$}. The Gibbs sampler is implemented via the Gumble-Max trick \cite{maddison2014sampling} as follows:
\begin{align}
\mbox{\footnotesize $\omega_m$} &\mbox{\footnotesize $\overset{\mbox{\tiny i.i.d}}{\sim} \mbox{Gumbel}(0,1),\, m \in \{1,.\,.\,.,M\},$} \label{eq99}\\ 
\mbox{\footnotesize $z_{v,a}$} &\mbox{\footnotesize $\sim \, \arg \max_m \mathbb{P}(\mathcal{H}_t\,|\,z,z_{v,a} = m)(|\mathcal{A}_v^{(m)}|+\gamma_{m}) + \omega_m.$} \nonumber 
\end{align}  

\subsubsection{Exploration via Diverse Batch Selection} 
\label{BatchSec}
The BO procedure solves the PSCP problem by exploring the performances of a sequence of pipelines \mbox{\footnotesize $\{P^1_{\theta^1},P^2_{\theta^2},.\,.\,.\}$} until it (hopefully) converges to the optimal pipeline \mbox{\footnotesize $P^*_{\theta^*}$}. In every iteration \mbox{\footnotesize $t$}, BO picks a pipeline to evaluate using an {\it acquisition function} \mbox{\footnotesize $A(P_\theta; \mathcal{H}_{t})$} that balances between {\it exploration} and {\it exploitation}. \textsc{AutoPrognosis} deploys a 2-step batched (parallelized) exploration scheme that picks \mbox{\footnotesize $B$} pipelines for evaluation at every iteration \mbox{\footnotesize $t$} as follows:   

\Ovalbox{
\begin{minipage}{3.1in}
{\footnotesize {\bf \underline{Step 1:}} Select the frequentist kernel decomposition \mbox{\footnotesize $\{{\bf \hat{\Lambda}}^{(m)}\}_m$} that maximizes the posterior \mbox{\footnotesize $\mathbb{P}(z\,|\,\mathcal{H}_t)$.}}
\end{minipage}} 

\Ovalbox{
\begin{minipage}{3.1in}
{\footnotesize {\bf \underline{Step 2:}} Select the $B$ pipelines $\{P^{b}_\theta\}^B_{b=1}$ with the highest values for the acquisition function $\{A(P^b_\theta; \mathcal{H}_{t})\}^B_{b=1}$, such that each pipeline $P^{b}_\theta,\, b \in \{1,.\,.\,.,B\},$ involves a {\bf distinct prediction} algorithm from a {\bf distinct subspace} in \mbox{\footnotesize $\{{\bf \hat{\Lambda}}^{(m)}\}_m$}.}
\end{minipage}} 

We use the well-known {\it Upper Confidence Bound} (UCB) as acquisition function \cite{snoek2012practical}. The decomposition in (\ref{eq2}) offers an {\bf exponential speed up} in the overall {\bf computational complexity} of Step 2 since the UCB acquisition function is maximized separately for every (low-dimensional) component $f_m$; this reduces the number of computations from to \mbox{\footnotesize $\boldsymbol{\mathcal{O}(n^{-D})}$} to \mbox{\footnotesize $\boldsymbol{\mathcal{O}(n^{-D_m})}$}. The batched implementation is advantageous since sequential evaluations of \mbox{\footnotesize $f(.)$} are time consuming as it involves training the selected ML algorithms. 

Step 2 in the algorithm above encourages exploration as follows. In every iteration $t$, we select a ``diverse" batch of pipelines for which every pipeline is representative of a {\bf distinct} subspace in \mbox{\footnotesize $\{{\bf \hat{\Lambda}}^{(m)}\}_m$}. The batch selection scheme above encourages diverse exploration without the need for sampling pipelines via a determinantal point process with an exponential complexity as in \cite{kathuria2016batched,nikolov2015randomized,wang2017batched}. We also devise an efficient {\bf backward induction} algorithm that exploits the structure of a pipeline to maximize the acquisition function efficiently. (Details are provided in the supplement.)   

\section{Ensemble Construction \& Meta-learning}
\label{secccc3}
In this Section, we discuss the details of the ensemble Construction and meta-learning modules; details of the interpreter module are provided in the next Section. 
\subsection{Post-hoc Ensemble Construction}
The {\bf frequentist} approach to pipeline configuration is to pick the pipeline with the best observed performance from the set \mbox{\footnotesize $\{P^1_{\theta^1},.\,.\,.,P^t_{\theta^t}\}$} explored by the BO algorithm in Section \ref{BatchSec}. However, such an approach does not capture the uncertainty in the pipelines' performances, and wastefully throws away \mbox{\footnotesize $t-1$} of the evaluated pipelines. On the contrary, {\footnotesize \textsc{AutoPrognosis}} makes use of all such pipelines via post-hoc {\bf Bayesian model averaging}, where it creates an ensemble of weighted pipelines \mbox{\footnotesize $\sum_i w_i P^i_{\theta^i}$}. Model averaging is particularly useful in cohorts with small sample sizes, where large uncertainty about the pipelines' performances would render frequentist solutions unreliable. 

The ensemble weight \mbox{\footnotesize $w_i = \mathbb{P}(P^{i^*}_{\theta^{i^*}} = P^{i}_{\theta^{i}} \,|\, \mathcal{H}_t)$} is the posterior probability of \mbox{\footnotesize $P_\theta^i$} being the best performing pipeline: 
\begin{align}
\mbox{\small $w_i$} &\mbox{\small $= \sum_z \mathbb{P}(P^{i^*}_{\theta^{i^*}} = P^{i}_{\theta^{i}} \,|\, z, \mathcal{H}_t)\cdot \mathbb{P}(z \,|\, \mathcal{H}_t),$}
\label{erq99}
\end{align}    
where \mbox{\footnotesize $i^*$} is the pipeline configuration with the best (true) generalization performance. The weights in (\ref{erq99}) are computed by Monte Carlo sampling of kernel decompositions via the posterior \mbox{\footnotesize $\mathbb{P}(z \,|\, \mathcal{H}_t)$}, and then sampling the pipelines' performances from the posterior $f\,|\,z,\mathcal{H}_t$. Note that, unlike the ensemble builder of {\footnotesize \textsc{AutoSklearn}} \cite{feurer2015efficient}, the weights in (\ref{erq99}) account for correlations between different pipelines, and hence it penalizes combinations of ``similar" pipelines even if they are performing well. Moreover, our post-hoc approach allows building ensembles without requiring extra hyperparameters: in {\footnotesize \textsc{AutoWeka}}, ensemble construction requires a 5-fold increase in the number of hyperparameters \cite{kotthoff2016auto}.   

\subsection{Meta-learning via Empirical Bayes}
The Bayesian model used for solving the PSCP problem in Section \ref{sec3} can be summarized as follows:
\begin{align}
\mbox{\small $f \sim \mathcal{GP}(\mu, k\,|\,z),\,\, z \sim \mbox{Multi}(\alpha),\,\alpha \sim \mbox{Dirichlet}(M,\gamma).$} \nonumber
\end{align}

The speed of convergence of BO depends on the calibration of the prior's hyperparameters \mbox{\footnotesize $(M,\gamma,\mu,k)$}. An agnostic prior would require many iterations to converge to satisfactory pipeline configurations. To warmstart the BO procedure for a new cohort \mbox{\footnotesize $\mathcal{D}$}, we incorporate prior information obtained from previous runs of {\footnotesize \textsc{AutoPrognosis}} on a repository of \mbox{\footnotesize $K$} complementary cohorts \mbox{\footnotesize $\{\mathcal{D}_1,.\,.\,.,\mathcal{D}_K\}$}. Our meta-learning approach combines \mbox{\footnotesize $\{\mathcal{H}^1_{t_1},.\,.\,.,\mathcal{H}^M_{t_K}\}$} (optimizer runs on the \mbox{\footnotesize $K$} complementary cohorts) with the data in \mbox{\footnotesize $\mathcal{D}$} to obtain an empirical Bayes estimate \mbox{\footnotesize $(\hat{M},\hat{\gamma},\hat{\mu},\hat{k})$}.  

Our approach to meta-learning works as follows. For every complementary dataset \mbox{\footnotesize $\mathcal{D}_k$}, we create a set of 55 {\bf meta-features} \mbox{\footnotesize $\mathcal{M}(\mathcal{D}_k)$}, 40 of which are {\bf statistical meta-features} (e.g. number of features, size of data, class imbalance, etc), and the remaining 15 are {\bf clinical meta-features} (e.g. lab tests, vital signs, ICD-10 codes, diagnoses, etc). For every complementary dataset in \mbox{\footnotesize $\mathcal{D}_j$}, we optimize the hyperparameters \mbox{\footnotesize $(\hat{M}_j,\hat{\gamma}_j,\hat{\mu}_j,\hat{k}_j)$} via marginal likelihood maximization. For a new cohort \mbox{\footnotesize $\mathcal{D}$}, we compute a set of weights \mbox{\footnotesize $\{\eta_j\}_{j}$}, with \mbox{\footnotesize $\eta_j = \ell_j/\sum_k \ell_k,$} where \mbox{\footnotesize $\ell_j = \|\mathcal{M}(\mathcal{D})-\mathcal{M}(\mathcal{D}_j)\|_1$}, and calibrate its prior \mbox{\footnotesize $(M,\gamma,\mu,k)$} by setting it to be the average of the estimates \mbox{\footnotesize $(\hat{M}_j,\hat{\gamma}_j,\hat{\mu}_j,\hat{k}_j)$}, weighted by \mbox{\footnotesize $\{\eta_j\}_{j}$}. 

Existing methods for meta-learning focus only on identifying well-performing pipelines from other datasets, and use them for initializing the optimization procedure \cite{brazdil2008metalearning,feurer2015efficient}. Conceptualizing meta-learning as an empirical Bayes calibration procedure allows the transfer of a much richer set of information across datasets. Through the method described above, {\footnotesize \textsc{AutoPrognosis}} can import information on the smoothness of the black-box function (\mbox{\footnotesize $k$}), the similarities among baseline algorithms (\mbox{\footnotesize $\gamma,M$}), and the expected pipelines' performances (\mbox{\footnotesize $\mu$}). This improves not only the initialization of the BO procedure, but also the mechanism by which it explores the pipelines' design space. 
   
\begin{table*}[!htbp]
\centering
\begin{tcolorbox}[tab2,tabularx={l||cccc|cccccc}]           
 & {\tiny \bf MAGGIC} & {\tiny \bf UK Biobank} & {\tiny {\bf UNOS-I}} & {\tiny {\bf UNOS-II}} & {\tiny {\bf SEER-I}} & {\tiny {\bf SEER-II}} & {\tiny {\bf SEER-III}} & {\tiny {\bf SEER-IV}} & {\tiny {\bf SEER-V}} & {\tiny {\bf SEER-VI}} \\
\hline \hline
{\tiny \bf \textsc{AutoPrognosis}} & & & & & & & & & \\ 
{\tiny \,\,\,\,vanilla} & \mbox{\tiny 0.76 $\pm$ .004} & \mbox{\tiny 0.71 $\pm$ .004} & \mbox{\tiny 0.78 $\pm$ .002} & \mbox{\tiny 0.65 $\pm$ .001} & \mbox{\tiny 0.68 $\pm$ .002} & \mbox{\tiny 0.66 $\pm$ .005} & \mbox{\tiny 0.61 $\pm$ .001} & \mbox{\tiny 0.69 $\pm$ .002} & \mbox{\tiny 0.64 $\pm$ .002} & \mbox{\tiny 0.65 $\pm$ .003} \\ 
{\tiny \,\,\,\,\,\,best predictor} & \mbox{\tiny \bf \textcolor{red}{Grad. Boost}} & \mbox{\tiny \bf \textcolor{red}{XGBoost}} & \mbox{\tiny \bf \textcolor{red}{AdaBoost}} & \mbox{\tiny \bf \textcolor{red}{Rand. Forest}} & \mbox{\tiny \bf \textcolor{blue}{Cox PH}} & \mbox{\tiny \bf \textcolor{blue}{Cox PH}} & \mbox{\tiny \bf \textcolor{blue}{S. Forest}} & \mbox{\tiny \bf \textcolor{blue}{Cox PH}} & \mbox{\tiny \bf \textcolor{blue}{S. Forest}} & \mbox{\tiny \bf \textcolor{blue}{Cox PH}} \\  
{\tiny \,\,\,\,+ ensembles} & \mbox{\tiny 0.77 $\pm$ .002} & \mbox{\tiny 0.73 $\pm$ .003} & \mbox{\tiny 0.80 $\pm$ .001} & \mbox{\bf \tiny 0.66 $\pm$ .001} & \mbox{\tiny 0.68 $\pm$ .002} & \mbox{\tiny 0.67 $\pm$ .003} & \mbox{\tiny 0.62 $\pm$ .001} & \mbox{\tiny 0.69 $\pm$ .002} & \mbox{\tiny 0.66 $\pm$ .002} & \mbox{\tiny 0.65 $\pm$ .002} \\  
{\tiny \,\,\,\,+ meta-learning} & \mbox{\tiny 0.77 $\pm$ .004} & \mbox{\tiny 0.72 $\pm$ .004} & \mbox{\tiny 0.79 $\pm$ .002} & \mbox{ \tiny 0.65 $\pm$ .002} & \mbox{\tiny 0.72 $\pm$ .003} & \mbox{\tiny 0.68 $\pm$ .003} & \mbox{\tiny 0.64 $\pm$ .001} & \mbox{\tiny 0.71 $\pm$ .003} & \mbox{\tiny 0.69 $\pm$ .003} & \mbox{\tiny 0.66 $\pm$ .002} \\  
{\tiny \,\,\,\,full-fledged} & \mbox{\bf \tiny 0.78 $\pm$ .004} & \mbox{\bf \tiny 0.74 $\pm$ .003} & \mbox{\bf \tiny 0.81 $\pm$ .001} & \mbox{\bf \tiny 0.66 $\pm$ .001} & \mbox{\bf \tiny 0.73 $\pm$ .003} & \mbox{\bf \tiny 0.69 $\pm$ .003} & \mbox{\bf \tiny 0.64 $\pm$ .001} & \mbox{\bf \tiny 0.72 $\pm$ .002} & \mbox{\bf \tiny 0.70 $\pm$ .003} & \mbox{\bf \tiny 0.67 $\pm$ .002} \\ \hline \hline
{\tiny \textsc{Auto-sklearn}} & \mbox{\tiny 0.76 $\pm$ .003}& \mbox{\tiny 0.72 $\pm$ .004} & \mbox{\tiny 0.77 $\pm$ .002} & \mbox{\tiny 0.63 $\pm$ .002} & \mbox{\tiny 0.67 $\pm$ .002} & \mbox{\tiny 0.51 $\pm$ .005} & \mbox{\tiny 0.60 $\pm$ .001} & \mbox{\tiny 0.65 $\pm$ .004} & \mbox{\tiny 0.64 $\pm$ .002} & \mbox{\tiny 0.61 $\pm$ .003} \\ 
{\tiny \textsc{Auto-Weka}} & \mbox{\tiny 0.75 $\pm$ .003}& \mbox{\tiny 0.72 $\pm$ .005} & \mbox{\tiny 0.78 $\pm$ .001} & \mbox{\tiny 0.62 $\pm$ .002} & \mbox{\tiny 0.66 $\pm$ .002} & \mbox{\tiny 0.54 $\pm$ .004} & \mbox{\tiny 0.59 $\pm$ .002} & \mbox{\tiny 0.68 $\pm$ .003} & \mbox{\tiny 0.63 $\pm$ .004} & \mbox{\tiny 0.63 $\pm$ .002} \\ 
{\tiny \textsc{TPOT}} & \mbox{\tiny 0.74 $\pm$ .006} & \mbox{\tiny 0.68 $\pm$ .005} & \mbox{\tiny 0.72 $\pm$ .003} & \mbox{\tiny 0.61 $\pm$ .003} & \mbox{\tiny 0.64 $\pm$ .003} & \mbox{\tiny 0.59 $\pm$ .003} & \mbox{\tiny 0.57 $\pm$ .002} & \mbox{\tiny 0.67 $\pm$ .004} & \mbox{\tiny 0.62 $\pm$ .005} & \mbox{\tiny 0.61 $\pm$ .003} \\   \hline \hline
{\tiny Clinical Score} & \mbox{\tiny 0.70 $\pm$ .007} & \mbox{\tiny 0.70 $\pm$ .003} & \mbox{\tiny 0.62 $\pm$ .001} & \mbox{\tiny 0.56 $\pm$ .001} & \textemdash & \textemdash & \textemdash & \textemdash & \textemdash & \textemdash \\ \hline \hline
{\tiny Cox PH} & \mbox{\tiny 0.75 $\pm$ .005} & \mbox{\tiny 0.71 $\pm$ 0.002} & \mbox{\tiny 0.70 $\pm$ .001} & \mbox{\tiny 0.59 $\pm$ .001} & \mbox{\tiny 0.71 $\pm$ .003} & \mbox{\tiny 0.65 $\pm$ .004} & \mbox{\tiny 0.62 $\pm$ .002} & \mbox{\bf \tiny 0.72 $\pm$ .003} & \mbox{\tiny 0.68 $\pm$ .003} & \mbox{\bf \tiny \underline{0.67 $\pm$ .002}} \\  
\end{tcolorbox}
\caption{\footnotesize Performance of the different prognostic models in terms of the AUC-ROC with 5-fold cross-validation. Bold numbers correspond to the best result. The ``best predictor" row lists the prediction algorithms picked by vanilla \textsc{AutoPrognosis}.}
\label{Table2}
\end{table*}

\section{Evaluation of \textsc{AutoPrognosis}}
\label{sec4}
In~this~section,~we~assess~the~ability~of~{\footnotesize \textsc{AutoPrognosis}} to {\bf automatically} make the right prognostic {\bf modeling choices} when confronted with a variety of clinical datasets with different {\bf meta-features}. 

\subsection{Cardiovascular Disease Cohorts} 
We conducted experiments on 10 cardiovascular cohorts that correspond to the following aspects of patient care: 

{\bf $\bullet$ \textcolor{red}{\underline{Preventive care}}:} We considered two major cohorts for preventive cardiology. The first is the Meta-analysis Global Group in Chronic heart failure database ({\footnotesize {\bf MAGGIC}}), which holds data for 46,817 patients gathered from multiple clinical studies \cite{wong2014heart}. The second cohort is the {\footnotesize {\bf UK Biobank}}, which is a bio-repository with data for more than 500,000 volunteers in the UK \cite{sudlow2015uk}. 

{\bf $\bullet$ \textcolor{red}{\underline{Heart trans}p\underline{lant wait-list mana}g\underline{ement}}:} We extracted data from the United Network for Organ Sharing (UNOS) database, which holds information on all heart transplants conducted in the US between the years 1985 to 2015. Cohort {\footnotesize {\bf UNOS-I}} is a pre-transplant population of 36,329 cardiac patients who were enrolled in a transplant wait-list. 

{\bf $\bullet$ \textcolor{red}{\underline{Post-trans}p\underline{lant follow-u}p}:} Cohort {\footnotesize {\bf UNOS-II}} is a post-transplant population of 60,400 patients in the US who underwent a transplant between the years 1985 to 2015.  

{\bf $\bullet$ \textcolor{blue}{\underline{Cardiovascular comorbidities}}:} We extracted 6 cohorts from the Surveillance, Epidemiology, and End Results (SEER) cancer registries, which cover approximately 28$\%$ of the US population \cite{yoo2019surveillance}. We predict cardiac deaths in patients diagnosed with breast cancer ({\footnotesize {\bf SEER-I}}), colorectal cancer ({\footnotesize {\bf SEER-II}}), Leukemia ({\footnotesize {\bf SEER-III}}), respiratory cancers ({\footnotesize {\bf SEER-IV}}), digestive system cancer ({\footnotesize {\bf SEER-V}}), and urinary system cancer ({\footnotesize {\bf SEER-VI}}). 

The first three groups of datasets (colored in red) were collected for cohorts of patients diagnosed with (or at risk for) cardiac diseases, and so they shared a set of meta-features, including a \textcolor{red}{large number of cardiac risk factors}, \textcolor{red}{low censoring rate}, and \textcolor{red}{moderate class imbalance}. The last group of datasets (colored in blue) was collected for cohorts of cancer patients for whom cardiac diseases are potential comorbidities. These datasets shared a different set of meta-features, including a \textcolor{blue}{small number of cardiac risk factors}, \textcolor{blue}{high censoring rate}, and \textcolor{blue}{severe class imbalance}. Our experiments will demonstrate the ability of {\footnotesize \textsc{AutoPrognosis}} to adapt its modeling choices to these different clinical setups.    

\subsection{Performance of  {\footnotesize \textsc{AutoPrognosis}}}
\label{secA} 
Table \ref{Table2} shows the performance of various competing prognostic modeling approaches evaluated in terms of the area under receiver operating characteristic curve (AUC-ROC) with 5-fold cross-validation\footnote{All algorithms were allowed to run for a maximum of 10 hours to ensure a fair comparison.}. We compared the performance of {\footnotesize \textsc{AutoPrognosis}} with the clinical risk scores used for predicting prognosis in each cohort (MAGGIC score in {\footnotesize {\bf MAGGIC}} and {\footnotesize {\bf UNOS-I}} \cite{wong2014heart}, Framingham score in the {\footnotesize {\bf UK Biobank}} \cite{schnabel2009development}, and IMPACT score in {\footnotesize {\bf UNOS-II}} \cite{weiss2011creation}). We also compared with various AutoML frameworks, including {\footnotesize \textsc{Auto-Weka}} \cite{kotthoff2016auto}, {\footnotesize \textsc{Auto-sklearn}} \cite{feurer2015efficient}, and {\footnotesize \textsc{TPOT}} \cite{olson2016tpot}. Finally, we compared with a standard Cox proportional hazards (Cox PH) model, which is the model most commonly used in clinical prognostic research. 

\begin{figure}[h]
\centering
\includegraphics[width=3in]{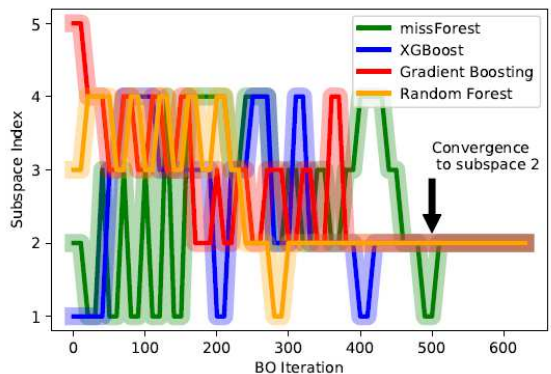}
\caption{\footnotesize The learned kernel decomposition for {\footnotesize {\bf MAGGIC}}.}
\label{fig55}
\end{figure}

Table \ref{Table2} demonstrates the superiority of {\footnotesize \textsc{AutoPrognosis}} to all the competing models on all the cohorts under consideration. This reflects the robustness of our system since the 10 cohorts had very different characteristics. In many experiments, the learned kernel decomposition reflected an intuitive clustering of algorithms by the similarity of their structure. For instance, Figure \ref{fig55} shows one subspace in the frequentist decomposition learned by {\footnotesize \textsc{AutoPrognosis}} over the BO iterations for the {\footnotesize {\bf MAGGIC}} cohorts. We can see that all ensemble methods in the imputation and prediction stages that use decision-trees as their base learners were lumped together in the same subspace.

\subsection{The ``Interpreter"}
\label{secC}
Albeit accurate, models built by {\footnotesize \textsc{AutoPrognosis}} would generally be hard for a clinician to ``interpret". To address this issue, {\footnotesize \textsc{AutoPrognosis}} deploys an {\bf interpreter} module (see Figure \ref{fig1}) that takes as an input the learned model for a given cohort, in addition to a set of actionable risk strata \mbox{\footnotesize $\mathcal{R}$}, and outputs an ``explanation" for its predictions in terms of a set of logical {\it association rules} of the form:
\begin{align}
\mbox{\footnotesize $C_1 \land C_2 \land .\,.\,. \land C_{l(r)} \implies r,\, \forall r \in \mathcal{R},$}
\label{eqqq1}
\end{align}  
where \mbox{\footnotesize $\{C_1,.\,.\,.,C_{l(r)}\}$} is a set of Boolean conditions associated with risk stratum $r$. The association rules are obtained via a {\it Bayesian associative classifier} \cite{ma1998integrating,agrawal1993mining,kruschke2008bayesian,luo2016automatically}, with a prior over association rules, and a posterior computed based on target labels that correspond to the outputs of the learned model discretized via the strata in \mbox{\footnotesize $\mathcal{R}$}. The Bayesian approach allows incorporating prior knowledge (from clinical literature) about ``likely" association rules.    

We report one example for an explanation provided by the interpreter module based on our experiments on the {\footnotesize \bf UK Biobank} cohort. For this cohort, the standard Framingham risk score exhibited an AUC-ROC of 0.705 for the overall cohort, but its AUC-ROC for patients with Type-2 Diabetes (T2D) was as low as 0.63. On the contrary, {\footnotesize \textsc{AutoPrognosis}} performed almost equally well in the two subgroups. The interpreter provided an explanation for the improved predictions through the following association rule: 

\Ovalbox{
\begin{minipage}{3.1in}
$\mbox{\footnotesize {\bf Diabetic}} \wedge \mbox{\footnotesize {\bf Lipid-lowering}} \wedge \mbox{\footnotesize {\bf (Age $\geq$ 40)}} \implies \mbox{\footnotesize {\bf High risk}}$ 
\end{minipage}}  

None of these risk factors were included in the standard guidelines. That is, the interpreter indicates that a better stratification, with new risk factors such the usage of lipid-lowering drugs, is possible for diabetic patients. Clinicians can use the interpreter as a data-driven hypothesis generator that prompts new risk factors and strata for subsequent research.

\subsection{Learning to Pick the Right Model and {\footnotesize \textsc{AutoPrognosis}} as a Clairvoyant}
We split up Table \ref{Table2} into 2 groups of columns: \textcolor{red}{group 1} (left) contains cohorts obtained from \textcolor{red}{cardiology studies}, whereas \textcolor{blue}{group 2} (right) contains cohorts obtained from \textcolor{blue}{cancer studies}, with cardiac secondary outcomes. As mentioned earlier, the two groups had different meta-features. We tracked the modeling choices made by vanilla {\footnotesize \textsc{AutoPrognosis}} (no ensembles or meta-learning) in both groups (``best predictor" row in Table \ref{Table2}). For all datasets in \textcolor{blue}{group 2}, {\footnotesize \textsc{AutoPrognosis}} decided that survival modeling (using Cox PH model or survival forests) is the right model. This is because, with the high prevalence of censored time-to-event data, survival models are more data-efficient than operating on binarized survival labels and removing patients lost to follow-up. When given richer datasets with a large number of relevant features, low rates of censoring and moderate imbalance (\textcolor{red}{group 1}), {\footnotesize \textsc{AutoPrognosis}} spent more iterations navigating ML classifiers, and learned that an algorithm like AdaBoost is a better choice for a dataset like {\footnotesize \bf UNOS-I}. Such a (non-intuitive) choice would have not been possibly identified by a clinical researcher; researchers typically use the Cox PH model, which on the {\footnotesize \bf UNOS-I} cohort provides an inferior performance.  
   
{\bf Meta-learning} was implemented via leave-one-dataset-out validation: we run vanilla {\footnotesize \textsc{AutoPrognosis}} on all of the 10 cohorts, and then for every cohort, we use the other 9 cohorts as the complementary datasets used to implement the meta-learning algorithm. Since the pool of complementary cohorts contained 5 datasets for cardiovascular comorbidities, meta-learning was most useful for \textcolor{blue}{group 2} datasets as they all had very similar meta-features. With meta-learning, {\footnotesize \textsc{AutoPrognosis}} had a strong prior on survival models for \textcolor{blue}{group 2} datasets, and hence it converges quickly to a decision on using a survival model having observed the dataset's meta-features. Ensemble construction was most useful for the {\footnotesize {\bf MAGGIC}} and {\footnotesize {\bf UNOS}} cohorts, since those datasets had more complex hypotheses to learn.   

Clinical researchers often ask the question: {\bf when should I use machine learning for my prognostic study?} The answer depends on the nature of the dataset involved. As we have see in Table \ref{Table2}, a simple Cox model may in some cases be sufficient to issue accurate predictions. The meta-learning module in {\footnotesize \textsc{AutoPrognosis}} can act as a clairvoyant that tells whether ML models would add value to a given prognostic study {\bf without even training any model}. That is, by looking at the ``meta-learned" GP prior calibrated by a new dataset's meta-features, we can see whether the prior assigns high scores to ML models compared to a simple Cox model, and hence decide on whether ML has gains to offer for such a dataset. 
  
\bibliography{icml20181}

\begin{thebibliography}{40}
\providecommand{\natexlab}[1]{#1}
\providecommand{\url}[1]{\texttt{#1}}
\expandafter\ifx\csname urlstyle\endcsname\relax
  \providecommand{\doi}[1]{doi: #1}\else
  \providecommand{\doi}{doi: \begingroup \urlstyle{rm}\Url}\fi

\bibitem[Agrawal et~al.(1993)Agrawal, Imieli{\'n}ski, and
  Swami]{agrawal1993mining}
Agrawal, Rakesh, Imieli{\'n}ski, Tomasz, and Swami, Arun.
\newblock Mining association rules between sets of items in large databases.
\newblock In \emph{Acm sigmod record}, volume~22, pp.\  207--216. ACM, 1993.

\bibitem[Alaa et~al.(2017)Alaa, Hu, and van~der Schaar]{alaa2017learning}
Alaa, Ahmed~M, Hu, Scott, and van~der Schaar, Mihaela.
\newblock Learning from clinical judgments: Semi-markov-modulated marked hawkes
  processes for risk prognosis.
\newblock \emph{International Conference on Machine Learning}, 2017.

\bibitem[Bergstra et~al.(2011)Bergstra, Bardenet, K{\'e}gl, and
  Bengio]{bergstra2011implementations}
Bergstra, James, Bardenet, R{\'e}mi, K{\'e}gl, B, and Bengio, Y.
\newblock Implementations of algorithms for hyper-parameter optimization.
\newblock In \emph{NIPS Workshop on Bayesian optimization}, pp.\ ~29, 2011.

\bibitem[Blaha(2016)]{blaha2016critical}
Blaha, Michael~J.
\newblock The critical importance of risk score calibration, 2016.

\bibitem[Brazdil et~al.(2008)Brazdil, Carrier, Soares, and
  Vilalta]{brazdil2008metalearning}
Brazdil, Pavel, Carrier, Christophe~Giraud, Soares, Carlos, and Vilalta,
  Ricardo.
\newblock \emph{Metalearning: Applications to data mining}.
\newblock Springer Science \& Business Media, 2008.

\bibitem[Cabitza et~al.(2017)Cabitza, Rasoini, and
  Gensini]{cabitza2017unintended}
Cabitza, Federico, Rasoini, Raffaele, and Gensini, Gian~Franco.
\newblock Unintended consequences of machine learning in medicine.
\newblock \emph{Jama}, 318\penalty0 (6):\penalty0 517--518, 2017.

\bibitem[Chen \& Guestrin(2016)Chen and Guestrin]{chen2016xgboost}
Chen, Tianqi and Guestrin, Carlos.
\newblock Xgboost: A scalable tree boosting system.
\newblock In \emph{Proceedings of the 22nd acm sigkdd international conference
  on knowledge discovery and data mining}, pp.\  785--794. ACM, 2016.

\bibitem[Feurer et~al.(2015)Feurer, Klein, Eggensperger, Springenberg, Blum,
  and Hutter]{feurer2015efficient}
Feurer, Matthias, Klein, Aaron, Eggensperger, Katharina, Springenberg, Jost,
  Blum, Manuel, and Hutter, Frank.
\newblock Efficient and robust automated machine learning.
\newblock In \emph{Advances in Neural Information Processing Systems (NIPS)},
  pp.\  2962--2970, 2015.

\bibitem[Fine \& Gray(1999)Fine and Gray]{fine1999proportional}
Fine, Jason~P and Gray, Robert~J.
\newblock A proportional hazards model for the subdistribution of a competing
  risk.
\newblock \emph{Journal of the American statistical association}, 94\penalty0
  (446):\penalty0 496--509, 1999.

\bibitem[Golovin et~al.(2017)Golovin, Solnik, Moitra, Kochanski, Karro, and
  Sculley]{golovin2017google}
Golovin, Daniel, Solnik, Benjamin, Moitra, Subhodeep, Kochanski, Greg, Karro,
  John, and Sculley, D.
\newblock Google vizier: A service for black-box optimization.
\newblock In \emph{Proceedings of the 23rd ACM SIGKDD International Conference
  on Knowledge Discovery and Data Mining}, pp.\  1487--1495. ACM, 2017.

\bibitem[Gy{\"o}rfi et~al.(2006)Gy{\"o}rfi, Kohler, Krzyzak, and
  Walk]{gyorfi2006distribution}
Gy{\"o}rfi, L{\'a}szl{\'o}, Kohler, Michael, Krzyzak, Adam, and Walk, Harro.
\newblock \emph{A distribution-free theory of nonparametric regression}.
\newblock Springer Science \& Business Media, 2006.

\bibitem[Hripcsak et~al.(2015)Hripcsak, Albers, and
  Perotte]{hripcsak2015parameterizing}
Hripcsak, George, Albers, David~J, and Perotte, Adler.
\newblock Parameterizing time in electronic health record studies.
\newblock \emph{Journal of the American Medical Informatics Association},
  22\penalty0 (4):\penalty0 794--804, 2015.

\bibitem[Hutter et~al.(2009)Hutter, Hoos, Leyton-Brown, and
  St{\"u}tzle]{hutter2009paramils}
Hutter, Frank, Hoos, Holger~H, Leyton-Brown, Kevin, and St{\"u}tzle, Thomas.
\newblock Paramils: an automatic algorithm configuration framework.
\newblock \emph{Journal of Artificial Intelligence Research}, 36\penalty0
  (1):\penalty0 267--306, 2009.

\bibitem[Hutter et~al.(2011)Hutter, Hoos, and
  Leyton-Brown]{hutter2011sequential}
Hutter, Frank, Hoos, Holger~H, and Leyton-Brown, Kevin.
\newblock Sequential model-based optimization for general algorithm
  configuration.
\newblock \emph{LION}, 5:\penalty0 507--523, 2011.

\bibitem[Ishwaran et~al.(2008)Ishwaran, Kogalur, Blackstone, and
  Lauer]{ishwaran2008random}
Ishwaran, Hemant, Kogalur, Udaya~B, Blackstone, Eugene~H, and Lauer, Michael~S.
\newblock Random survival forests.
\newblock \emph{The annals of applied statistics}, pp.\  841--860, 2008.

\bibitem[Jenatton et~al.(2017)Jenatton, Archambeau, Gonz{\'a}lez, and
  Seeger]{jenatton2017bayesian}
Jenatton, Rodolphe, Archambeau, Cedric, Gonz{\'a}lez, Javier, and Seeger,
  Matthias.
\newblock Bayesian optimization with tree-structured dependencies.
\newblock In \emph{International Conference on Machine Learning}, pp.\
  1655--1664, 2017.

\bibitem[Kandasamy et~al.(2015)Kandasamy, Schneider, and
  P{\'o}czos]{kandasamy2015high}
Kandasamy, Kirthevasan, Schneider, Jeff, and P{\'o}czos, Barnab{\'a}s.
\newblock High dimensional bayesian optimisation and bandits via additive
  models.
\newblock In \emph{International Conference on Machine Learning (ICML)}, pp.\
  295--304, 2015.

\bibitem[Kathuria et~al.(2016)Kathuria, Deshpande, and
  Kohli]{kathuria2016batched}
Kathuria, Tarun, Deshpande, Amit, and Kohli, Pushmeet.
\newblock Batched gaussian process bandit optimization via determinantal point
  processes.
\newblock In \emph{Advances in Neural Information Processing Systems}, pp.\
  4206--4214, 2016.

\bibitem[Kotthoff et~al.(2016)Kotthoff, Thornton, Hoos, Hutter, and
  Leyton-Brown]{kotthoff2016auto}
Kotthoff, Lars, Thornton, Chris, Hoos, Holger~H, Hutter, Frank, and
  Leyton-Brown, Kevin.
\newblock Auto-weka 2.0: Automatic model selection and hyperparameter
  optimization in weka.
\newblock \emph{Journal of Machine Learning Research}, 17:\penalty0 1--5, 2016.

\bibitem[Kruschke(2008)]{kruschke2008bayesian}
Kruschke, John~K.
\newblock Bayesian approaches to associative learning: From passive to active
  learning.
\newblock \emph{Learning \& behavior}, 36\penalty0 (3):\penalty0 210--226,
  2008.

\bibitem[Luo(2016)]{luo2016automatically}
Luo, Gang.
\newblock Automatically explaining machine learning prediction results: a
  demonstration on type 2 diabetes risk prediction.
\newblock \emph{Health information science and systems}, 4\penalty0
  (1):\penalty0 2, 2016.

\bibitem[Luo et~al.(2017)Luo, Stone, Johnson, Tarczy-Hornoch, Wilcox, Mooney,
  Sheng, Haug, and Nkoy]{luo2017automating}
Luo, Gang, Stone, Bryan~L, Johnson, Michael~D, Tarczy-Hornoch, Peter, Wilcox,
  Adam~B, Mooney, Sean~D, Sheng, Xiaoming, Haug, Peter~J, and Nkoy, Flory~L.
\newblock Automating construction of machine learning models with clinical big
  data: proposal rationale and methods.
\newblock \emph{JMIR research protocols}, 6\penalty0 (8), 2017.

\bibitem[Ma \& Liu(1998)Ma and Liu]{ma1998integrating}
Ma, Bing Liu Wynne Hsu~Yiming and Liu, Bing.
\newblock Integrating classification and association rule mining.
\newblock In \emph{Proceedings of the fourth international conference on
  knowledge discovery and data mining}, 1998.

\bibitem[Maddison et~al.(2014)Maddison, Tarlow, and
  Minka]{maddison2014sampling}
Maddison, Chris~J, Tarlow, Daniel, and Minka, Tom.
\newblock A* sampling.
\newblock In \emph{Advances in Neural Information Processing Systems}, pp.\
  3086--3094, 2014.

\bibitem[Nikolov(2015)]{nikolov2015randomized}
Nikolov, Aleksandar.
\newblock Randomized rounding for the largest simplex problem.
\newblock In \emph{Proceedings of the forty-seventh annual ACM symposium on
  Theory of computing}, pp.\  861--870. ACM, 2015.

\bibitem[Olson \& Moore(2016)Olson and Moore]{olson2016tpot}
Olson, Randal~S and Moore, Jason~H.
\newblock Tpot: A tree-based pipeline optimization tool for automating machine
  learning.
\newblock In \emph{ICML Workshop on Automatic Machine Learning}, pp.\  66--74,
  2016.

\bibitem[Pedregosa et~al.(2011)Pedregosa, Varoquaux, Gramfort, Michel, Thirion,
  Grisel, Blondel, Prettenhofer, Weiss, Dubourg, et~al.]{pedregosa2011scikit}
Pedregosa, Fabian, Varoquaux, Ga{\"e}l, Gramfort, Alexandre, Michel, Vincent,
  Thirion, Bertrand, Grisel, Olivier, Blondel, Mathieu, Prettenhofer, Peter,
  Weiss, Ron, Dubourg, Vincent, et~al.
\newblock Scikit-learn: Machine learning in python.
\newblock \emph{Journal of Machine Learning Research}, 12\penalty0
  (Oct):\penalty0 2825--2830, 2011.

\bibitem[Raskutti et~al.(2009)Raskutti, Yu, and Wainwright]{raskutti2009lower}
Raskutti, Garvesh, Yu, Bin, and Wainwright, Martin~J.
\newblock Lower bounds on minimax rates for nonparametric regression with
  additive sparsity and smoothness.
\newblock In \emph{Advances in Neural Information Processing Systems}, pp.\
  1563--1570, 2009.

\bibitem[Rasmussen \& Williams(2006)Rasmussen and
  Williams]{rasmussen2006gaussian}
Rasmussen, Carl~Edward and Williams, Christopher~KI.
\newblock \emph{Gaussian processes for machine learning}, volume~1.
\newblock MIT press Cambridge, 2006.

\bibitem[Schnabel et~al.(2009)Schnabel, Sullivan, Levy, Pencina, Massaro,
  D'Agostino, Newton-Cheh, Yamamoto, Magnani, Tadros,
  et~al.]{schnabel2009development}
Schnabel, Renate~B, Sullivan, Lisa~M, Levy, Daniel, Pencina, Michael~J,
  Massaro, Joseph~M, D'Agostino, Ralph~B, Newton-Cheh, Christopher, Yamamoto,
  Jennifer~F, Magnani, Jared~W, Tadros, Thomas~M, et~al.
\newblock Development of a risk score for atrial fibrillation (framingham heart
  study): a community-based cohort study.
\newblock \emph{The Lancet}, 373\penalty0 (9665):\penalty0 739--745, 2009.

\bibitem[Snoek et~al.(2012)Snoek, Larochelle, and Adams]{snoek2012practical}
Snoek, Jasper, Larochelle, Hugo, and Adams, Ryan~P.
\newblock Practical bayesian optimization of machine learning algorithms.
\newblock In \emph{Advances in neural information processing systems (NIPS)},
  pp.\  2951--2959, 2012.

\bibitem[Sudlow et~al.(2015)Sudlow, Gallacher, Allen, Beral, Burton, Danesh,
  Downey, Elliott, Green, Landray, et~al.]{sudlow2015uk}
Sudlow, Cathie, Gallacher, John, Allen, Naomi, Beral, Valerie, Burton, Paul,
  Danesh, John, Downey, Paul, Elliott, Paul, Green, Jane, Landray, Martin,
  et~al.
\newblock Uk biobank: an open access resource for identifying the causes of a
  wide range of complex diseases of middle and old age.
\newblock \emph{PLoS medicine}, 12\penalty0 (3):\penalty0 e1001779, 2015.

\bibitem[Swersky et~al.(2014)Swersky, Duvenaud, Snoek, Hutter, and
  Osborne]{swersky2014raiders}
Swersky, Kevin, Duvenaud, David, Snoek, Jasper, Hutter, Frank, and Osborne,
  Michael~A.
\newblock Raiders of the lost architecture: Kernels for bayesian optimization
  in conditional parameter spaces.
\newblock \emph{arXiv preprint arXiv:1409.4011}, 2014.

\bibitem[Wang et~al.(2017)Wang, Li, Jegelka, and Kohli]{wang2017batched}
Wang, Zi, Li, Chengtao, Jegelka, Stefanie, and Kohli, Pushmeet.
\newblock Batched high-dimensional bayesian optimization via structural kernel
  learning.
\newblock \emph{International Conference on Machine Learning (ICML)}, 2017.

\bibitem[Wang et~al.(2013)Wang, Zoghi, Hutter, Matheson, De~Freitas,
  et~al.]{wang2013bayesian}
Wang, Ziyu, Zoghi, Masrour, Hutter, Frank, Matheson, David, De~Freitas, Nando,
  et~al.
\newblock Bayesian optimization in high dimensions via random embeddings.
\newblock In \emph{IJCAI}, pp.\  1778--1784, 2013.

\bibitem[Weiss et~al.(2011)Weiss, Allen, Arnaoutakis, George, Russell, Shah,
  and Conte]{weiss2011creation}
Weiss, Eric~S, Allen, Jeremiah~G, Arnaoutakis, George~J, George, Timothy~J,
  Russell, Stuart~D, Shah, Ashish~S, and Conte, John~V.
\newblock Creation of a quantitative recipient risk index for mortality
  prediction after cardiac transplantation (impact).
\newblock \emph{The Annals of thoracic surgery}, 92\penalty0 (3):\penalty0
  914--922, 2011.

\bibitem[Wong et~al.(2014)Wong, Hawkins, Petrie, Jhund, Gardner, Ariti, Poppe,
  Earle, Whalley, Squire, et~al.]{wong2014heart}
Wong, Chih~M, Hawkins, Nathaniel~M, Petrie, Mark~C, Jhund, Pardeep~S, Gardner,
  Roy~S, Ariti, Cono~A, Poppe, Katrina~K, Earle, Nikki, Whalley, Gillian~A,
  Squire, Iain~B, et~al.
\newblock Heart failure in younger patients: the meta-analysis global group in
  chronic heart failure (maggic).
\newblock \emph{European heart journal}, 35\penalty0 (39):\penalty0 2714--2721,
  2014.

\bibitem[Yang et~al.(2015)Yang, Tokdar, et~al.]{yang2015minimax}
Yang, Yun, Tokdar, Surya~T, et~al.
\newblock Minimax-optimal nonparametric regression in high dimensions.
\newblock \emph{The Annals of Statistics}, 43\penalty0 (2):\penalty0 652--674,
  2015.

\bibitem[Yoo \& Coughlin(2018)Yoo and Coughlin]{yoo2019surveillance}
Yoo, Wonsuk and Coughlin, Steven~S.
\newblock Surveillance, epidemiology, and end results (seer) data for
  monitoring cancer trends.
\newblock \emph{Journal of the Georgia Public Health Association}, 2018.

\bibitem[Yoon et~al.(2017)Yoon, Alaa, Cadeiras, and van~der
  Schaar]{yoon2017personalized}
Yoon, Jinsung, Alaa, Ahmed~M, Cadeiras, Martin, and van~der Schaar, Mihaela.
\newblock Personalized donor-recipient matching for organ transplantation.
\newblock In \emph{AAAI}, pp.\  1647--1654, 2017.

\end{thebibliography}
\bibliographystyle{icml2018}


\end{document}